\documentclass{article}
\usepackage{spconf,amsmath,graphicx}

\def\etal{{\textit{et al.}}}
\title{An efficient CNN for spectral reconstruction from RGB images}
\name{Yigit Baran Can, Radu Timofte\sthanks{Work supported by ETH Zurich and an NVIDIA hardware grant.}}
\address{Computer Vision Lab, ETH Zurich, Switzerland}
\begin{document}
%
\maketitle
\begin{abstract}
Recently, the example-based single image spectral reconstruction from RGB images task \textit{aka} spectral super-resolution was approached by means of deep learning by Galliani~\etal~\cite{DBLP:journals/corr/GallianiLMBS17}. The proposed very deep convolutional neural network (CNN) achieved superior performance on recent large benchmarks. However, Aeschbacher~\etal~\cite{Aeschbacher-ICCVW-2017} showed that comparable performance can be achieved by shallow learning method based on A+, a method introduced for image super-resolution by Timofte~\etal~\cite{DBLP:conf/accv/TimofteSG14-short}. In this paper, we propose a moderately deep CNN model and substantially improve the reported performance on three spectral reconstruction standard benchmarks: ICVL, CAVE, and NUS.
\end{abstract}
\section{Introduction}
\label{sec:introduction}

There is a wide variety and huge number of sources of visual data in today's world. The most common form of these visual data is the RGB images collected from standard cameras. The spectrum of the scene is mapped to three values, matching the human vision system's three cones system. However, capturing and analyzing a wider range of spectrum offers benefits. Medical applications utilize hyperspectral data extensively \cite{dicker2006differentiation,randeberg2006hyperspectral,stamatas2003hyperspectral} as well as segmentation tasks \cite{tarabalka2010segmentation,camps2014advances}. Remote sensing is another area that hyperspectral systems are used \cite{lillesand2014remote,borengasser2007hyperspectral,hege2004hyperspectral}. The problem is, however, capturing more spectral information and more spatial information create a trade-off. Most systems evolve to focus on spatial resolution, rather than hyperspectral information. 

The focus of this work is increasing the spectral resolution of a single RGB image by reconstructing channels/images for a desired set of wavelengths. In other words, given a coarse description of the spectrum of the scene, such as RGB image, infer the missing spectral information. This problem can be named as spectral reconstruction or spectral super resolution. The problem of spectral super resolution is under constrained since the aim is to estimate a larger number spectral bands (over 30) from generally much lower number of channels (usually the R,G,B channels). However, it has been shown that there is significant correlation between spectral bands~\cite{hyperspectral_remote}. This correlation can be used to infer the missing information. 

The complementary problem of spatial super resolution is extensively studied in literature~\cite{Nasrollahi2014,Agustsson_2017_CVPR_Workshops,Timofte_2017_CVPR_Workshops}. However, there are much less studies on spectral super resolution,  most recent ones of which are summarized in section~\ref{ssc:related_work}.

In this paper, a Convolutional Neural Network (CNN) based approach is proposed. Considering the limitations inherent to the problem such as lack of data and  differences among the response functions of hyperspectral sensors, it will be argued that a relatively shallow CNN can avoid overfitting and manage to learn the latent mapping from RGB images to the desired spectral resolution.  	

\vspace{-0.4cm}
\subsection{Related Work}
\label{ssc:related_work}
\vspace{-0.1cm}
The extensive study of the spatial super resolution problem in literature led to methods with impressive performances~\cite{DBLP:conf/accv/TimofteSG14-short,DBLP:conf/eccv/DongLHT14-short,DBLP:conf/cvpr/KimLL16a-short,Ledig_2017_CVPR,Timofte_2017_CVPR_Workshops}. The current state-of-the-art methods are CNN-based. In comparison, the complementary spectral reconstruction problem attracted much less attention. One proposed method tries to find the illumination conditions and uses radial basis functions to model the correspondence between RGB values of the given image and the reflectance of the objects in the scene~\cite{DBLP:conf/eccv/NguyenPB14-short}. Other approaches focus on sparse dictionary learning. Arad~\etal~\cite{DBLP:conf/eccv/AradB16-short} learn a sparse dictionary method with K-SVD and OMP, they also introduce ICVL with 201 image pairs, the largest dataset for spectral reconstruction to date. Recently, Galliani~\etal~\cite{DBLP:journals/corr/GallianiLMBS17} propose a CNN architecture based on Tiramisu network~\cite{DBLP:conf/cvpr/JegouDVRB17-short} to map the RGB values to hyperspectral bands. Their (very) deep CNN has 56 layers and learns an end-to-end mapping. The input patch is downscaled by max pooling through several layers and the upscaled back to original size through some layers of sub pixel upsampling. Very recently Aeschbacher~\etal~\cite{Aeschbacher-ICCVW-2017} proposed an A+ based method. They build upon the A+ method of Timofte~\etal~\cite{DBLP:conf/accv/TimofteSG14-short}, originally introduced for the spatial super resolution problem. Aeschbacher~\etal proposes a sparse dictionary representation based method. 
It operates directly on pixel values and trains dictionaries using K-SVD and OMP.  Offline anchored regressors are learned from the training samples, to map the low spectral resolution space (RGB) to the higher spectral resolution. Aeschbacher~\etal also reimplemented and improved the performance of the method of Arad~\etal. In this paper, we compare with the state-of-the-art methods: Galliani's deep CNN, Aeschbacher's A+ based method and reimplementation of Arad's method.   

\section{Proposed Method}
\label{sec:pagestyle}

Spectral reconstruction from an input RGB image is heavily ill-posed. Moreover, the existing datasets and train data are relatively small in size when compared with those available in the related problem of spatial super resolution. 
We place ourselves in between the shallow A+ method of Aeschabacher~\etal~\cite{Aeschbacher-ICCVW-2017} and the (very) deep Tiramisu-based method of Galliani~\etal~\cite{DBLP:journals/corr/GallianiLMBS17} and avoid overfitting to the training data with a novel moderately deep, rather shallow CNN with residual blocks.

\begin{figure}[h]
\centering
\begin{tabular}{cc}
\includegraphics[width=0.55\linewidth]{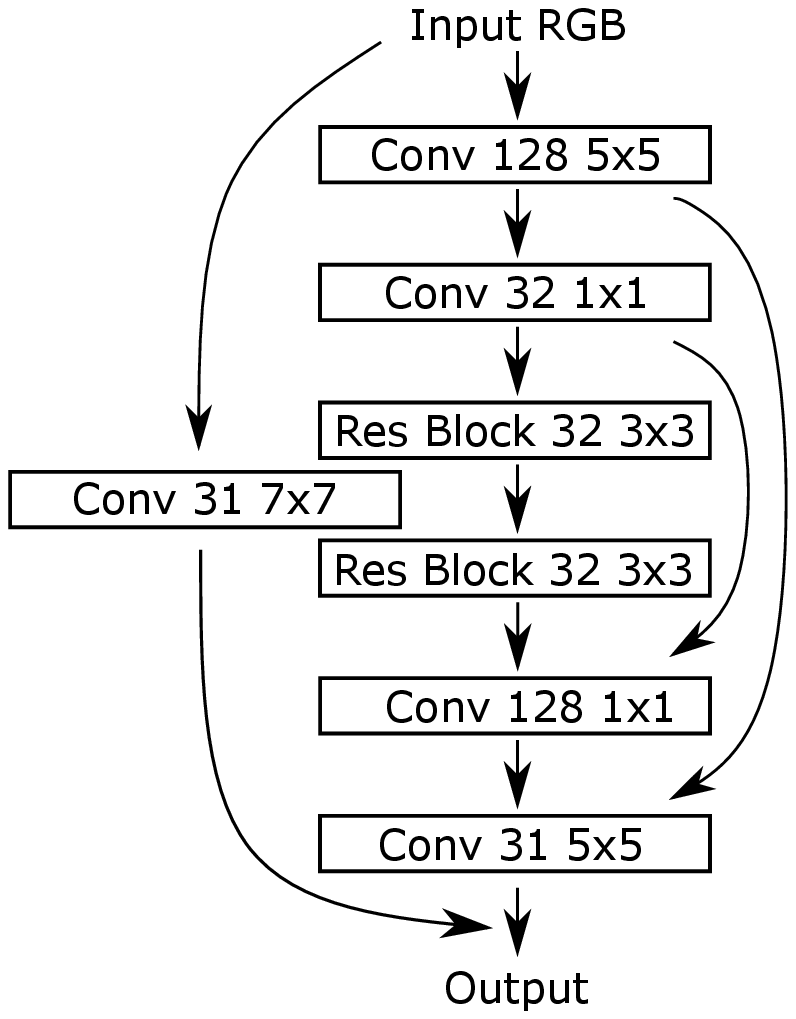}&
\includegraphics[width=0.3\linewidth]{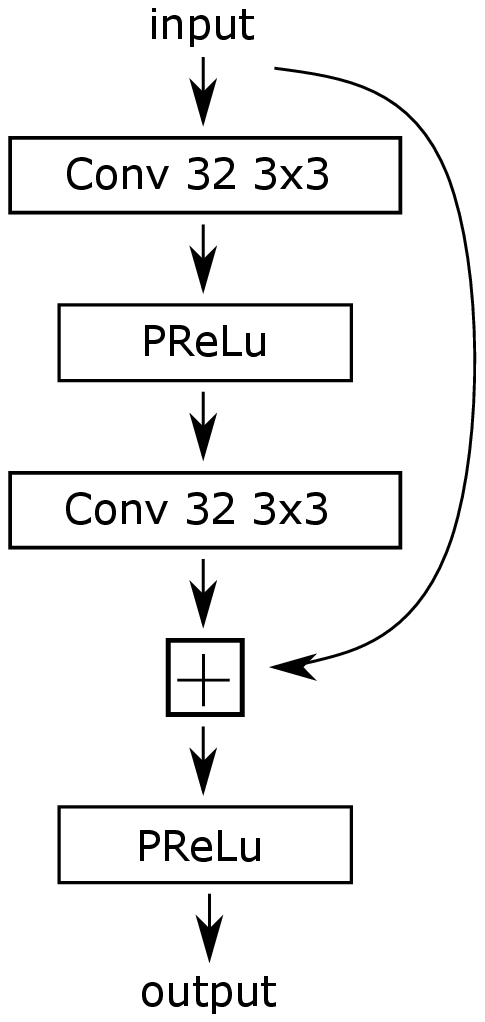}\\
a) network & b) res. block\\
\end{tabular}
\vspace{-0.1cm}
\caption{The proposed CNN architecture.}
\label{fig:proposed_architecture}
\vspace{-0.2cm}
\end{figure}

Figure~\ref{fig:proposed_architecture} gives the schematic representation of the proposed CNN network and the building residual block. `Conv' refers to convolutional layer, the number next to it refers to the number of feature maps of filters in that layer and the next element refers to the filter size. Arrows show the direction of flow. Wherever, the arrow heads meet the results of the layers at the source of the arrows are summed element-wise. The results of all layers except the last layer are passed through a PReLU, the formula of which is~\eqref{eq:PReLU}, as the non-linearity. PReLU~\cite{DBLP:conf/iccv/HeZRS15-short} was showed to improve over the traditional non-parametric ReLU.
\begin{equation}\label{eq:PReLU}
    f(y_i)= 
\begin{cases}
    y_i,		& \text{if } y_i > 0\\
    (a_i)(y_i), & \text{if }  y_i\le 0
\end{cases}
\end{equation}

The proposed architecture can be considered as two networks: the main network and the $7\times7$ conv layer. The architecture is created to form residual blocks all along the network. The $7\times 7$ conv layer can be considered as a skip connection of the residual block while the main network is core of the residual block. The $7\times7$ conv layer estimates the basic mapping from RGB to hyperspectral. For standard spatial super resolution, estimating the difference of high resolution image and the bicubic upsampled low resolution image is a common practice~\cite{Timofte_2017_CVPR_Workshops}. This convolution layer basically implements this operation but instead of a hand crafted method it learns the upsampling mapping. In the main network, the sub-network formed by layers from 2nd to 6th can be regarded as the residual block of the last layer and so on. 

Apart from the idea of forming residual sub-networks, we use regular residual blocks in the network. As shown in Figure~\ref{fig:proposed_architecture}, we opted for 2 residual blocks. Increasing the number of residual blocks brings only small benefits at the expense of runtime and potentially can lead to overfitting especially in our settings, some hyperspectral datasets have a small number of samples.

The initial features extracted from the input are shrunk with $1\times1$ conv layers to form a bottleneck and decrease the computation cost and time. The bottleneck decreases overfit and forces the network to learn more compact and relevant features. However, the pre-shrink features are utilized further in the network through the skip connections. Therefore, the source of the learned complex features is also used. This idea aligns with the main concept of the network which is forming residual parts all along the network. Generally, the initial layers of CNNs are responsible for learning simple features such as gradients or blobs. Although combining them to form more complex features and using them to make decisions in further segments of the network is beneficial, the simple features can also be useful in the further stages. The shrunken features are then processed by the residual blocks. The blocks are composed of $3\times3$ convolutional filters just like the original Residual Network paper~\cite{he2016deep}. Different than the original block, we have PReLU as activation function. 

\begin{figure*}[htb]
  \includegraphics[width=\linewidth]{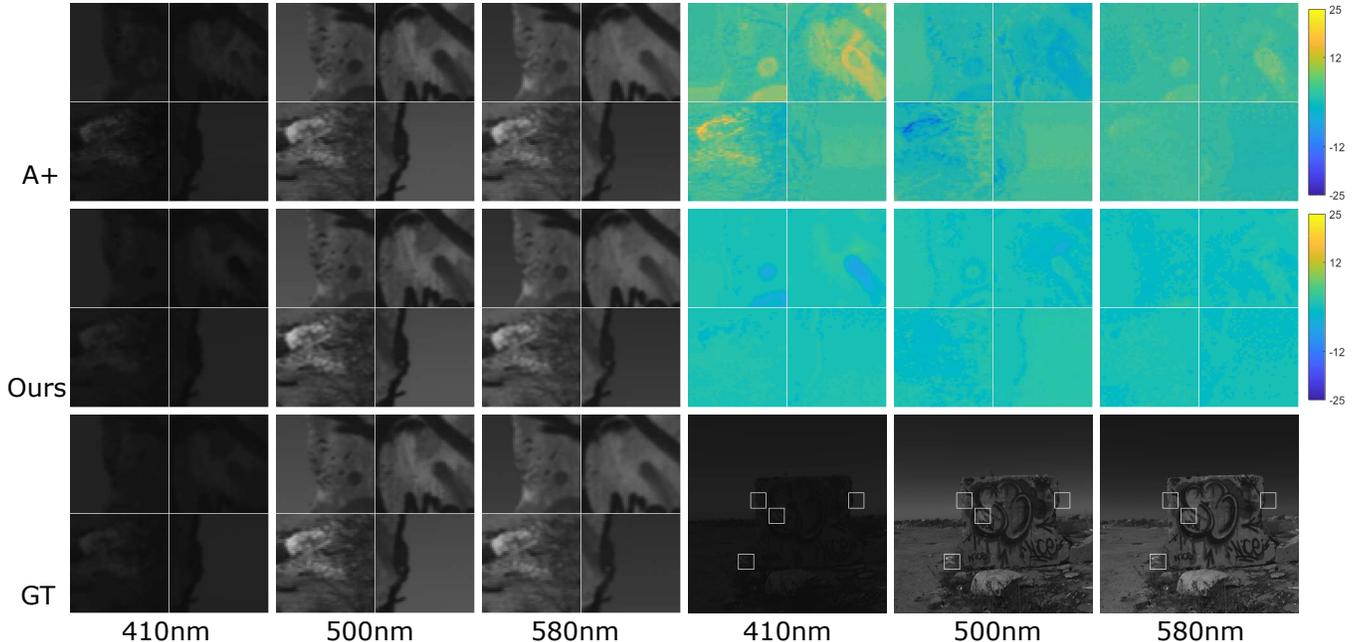}
  \vspace{-0.4cm}
  \caption{Visual comparison on ICVL between ground truth (GT) and reconstructed bands by A+~\cite{Aeschbacher-ICCVW-2017} and our method.}
  \label{fig:visual_comparison}
  \vspace{-0.3cm}
\end{figure*}

The output of the residual blocks are expanded from 32 back to the original feature map count of 128. This ensures that more features are available to the final layer. Since the bulk of the processing has past, this expansion does not increase computation time heavily. This layer can be seen as counterpart of the second layer where the initial features were shrunk. After the expansion of the output of the residual block, the resulting maps are passed to the final layer of $5\times5$ convolution. There are 31 maps in this layer corresponding to the 31 channel images we are reconstructing. The spatial extent of this layer is kept high to ensure that the nearby pixels are also taken into consideration. Finally, the result from the $7\times7$ convolution layer is added to form the output. 

The network's receptive field is $17\times17$. This means a pixel in the output is calculated using a local neighborhood of $17\times17$ pixels. With the $7\times7$ convolution layer, the local neighborhood of $7\times7$ pixels have an extra effect on the resulting pixel.

\vspace{-0.1cm}
\section{Experimental results}
\label{sec:experiments}

We compare our proposed approach against the methods from Galliani~\etal~\cite{DBLP:journals/corr/GallianiLMBS17} and Aeschbacher~\etal~\cite{Aeschbacher-ICCVW-2017} as roughly described in Section~\ref{sec:introduction}. We adhere to the experimental setup from~\cite{Aeschbacher-ICCVW-2017,DBLP:journals/corr/GallianiLMBS17} and report RMSE and relative RMSE (rRMSE) as defined in~\cite{Aeschbacher-ICCVW-2017} for 3 benchmarks: ICVL~\cite{DBLP:conf/eccv/AradB16-short}, CAVE~\cite{DBLP:journals/tip/YasumaMIN10}, and NUS~\cite{DBLP:conf/eccv/NguyenPB14-short}.
Because Galliani measured the errors using 8bit images, we report also our results \textit{w.r.t.} 8 bit images.

In NUS dataset we use the provided train/test split and for ICVL and CAVE we apply a 2 fold validation by dividing the images into two sets and training 2 models for each set. On each test RGB image we employ the model that did not use it in training. The results are averaged to give the final test error.

\vspace{-0.1cm}
\subsection{Datasets}
\label{ssec:subhead}



\noindent\textbf{ICVL} dataset of Arad~\etal~\cite{DBLP:conf/eccv/AradB16-short} includes 201 hyperspectral images with $1392\times1300$ resolution over 519 spectral bands (400-1,000nm). The images were captured by a line scanner camera (Specim PS Kappa DX4 hyperspectral). Although there are 519 bands, we used the downsampled version which has 31 bands from 400nm to 700nm with 10nm increments. Following the practice of Galliani and Aeschbacher we use the CIE 1964 color matching functions to prepare the corresponding RGB images of the hyperspectral images. 

\begin{table*}[th!]
\caption{Quantitative comparison on ICVL~\cite{DBLP:conf/eccv/AradB16-short}, CAVE~\cite{DBLP:journals/tip/YasumaMIN10} and NUS~\cite{DBLP:conf/eccv/NguyenPB14-short} datasets. Best results are in bold.}
\label{tab:MethodComparison}
\centering
\vspace{0.1cm}
\resizebox{\linewidth}{!}
{
\begin{tabular}{l||ccccc|ccccc|cccccc}
&\multicolumn{5}{|c|}{\textbf{ICVL dataset~\cite{DBLP:conf/eccv/AradB16-short}}}&\multicolumn{5}{|c|}{\textbf{CAVE dataset~\cite{DBLP:journals/tip/YasumaMIN10}}}&\multicolumn{6}{|c}{\textbf{NUS dataset~\cite{DBLP:conf/eccv/NguyenPB14-short}}} \\
 & Galliani  & \textbf{Arad}  & \textbf{A+} & \textbf{ours}& \textbf{ours+E} & Galliani   &Arad & \textbf{A+} & \textbf{ours} & \textbf{ours+E} & Nguyen & Galliani & Arad & \textbf{A+} & \textbf{ours} & \textbf{ours+E} \\
 & \cite{DBLP:journals/corr/GallianiLMBS17}& \cite{Aeschbacher-ICCVW-2017} &\cite{Aeschbacher-ICCVW-2017}&  & &\cite{DBLP:journals/corr/GallianiLMBS17}& \cite{Aeschbacher-ICCVW-2017} &\cite{Aeschbacher-ICCVW-2017}&  & &\cite{DBLP:conf/eccv/NguyenPB14-short}  & \cite{DBLP:journals/corr/GallianiLMBS17}& \cite{Aeschbacher-ICCVW-2017} &\cite{Aeschbacher-ICCVW-2017}&  & \\
\hline
\small{rRMSE} & -& 0.0507& 0.0344& 0.0168& \textbf{0.0166} &-  &0.4998 & 0.4265 & 0.4697& \textbf{0.178} & 0.2145 & -& 0.1904& \textbf{0.1420}&0.1524&0.1471  \\

\small{rRMSE$_G$} & -& 0.0873& 0.0584& 0.0401& \textbf{0.0399} &- & 0.7755& 0.3034 &0.246 &\textbf{0.239} & 0.3026&- & 0.3633 & 0.2242 & 0.2317& \textbf{0.2168}\\

\small{rRMSE$_G^{uint}$} & 0.0587& -& - & 0.0353& \textbf{0.0350}& 0.2804 & -& - &0.1525&\textbf{0.1482} & 0.3026&0.234 & -&- & 0.1796& \textbf{0.1747}\\

\small{RMSE}& -& 1.70& 1.04& 0.6407& \textbf{0.6324} & -& 5.61&2.74&\textbf{2.550}&2.613 &12.44& -& 4.44& 2.92& 2.86& \textbf{2.83}\\

\small{RMSE$_G$} & -& 3.24& 1.96& 1.35& \textbf{1.33}&- & 20.13& 6.70&\textbf{5.77}&5.80& 8.06& -& 9.56 & 5.17&5.12&\textbf{4.92}\\

\small{RMSE$_G^{uint}$} & 1.98& -& -& 1.25& \textbf{1.23}&4.76 & -& -&\textbf{3.4924}&3.5275& 8.06& 5.27& - & -&\textbf{3.66}&\textbf{3.66}\\

\end{tabular}
}
\vspace{-0.3cm}
\end{table*}

\noindent\textbf{CAVE} database proposed by Yasuma~\etal~\cite{DBLP:journals/tip/YasumaMIN10} has 32 images with $512\times512$ resolution. There are 31 bands for each image, ranging from 400 to 700 nm with 10 nm increments. The pictures were taken with a cooled CCD camera (Apogee Alta U260). The dataset contains various objects including food, fabric, faces and paints. In this dataset, Aeschbacher~\etal~\cite{Aeschbacher-ICCVW-2017} followed a 4 fold cross validation.

\noindent\textbf{NUS} dataset introduced by Nguyen~\etal~\cite{DBLP:conf/eccv/NguyenPB14-short}    contains 66 spectral images and their corresponding illuminations. Just like the other 2 datasets, the spectral bands range between 400 to 700nm, with 10nm increments. The pictures were taken with a Specim’s PFDCL- 65-V10E spectral camera. Different illumination conditions were used. Natural daylight and metal halide bands were utilized to form a diverse set. Here, following Galliani and Aeschbacher, instead of CIE 1964 mapping, Canon 1D Mark III response function was used to map the hyperspectral images to RGB space.


\vspace{-0.1cm}
\subsection{Implementation Details}
\label{ssc:implementation_details}

\noindent{\textbf{Training }}
The proposed network was trained with TensorFlow from scratch with Adam optimizer. The learning rate was initially set to 0.0005 while multiplied by 0.93 at every 50000 iterations. The networks were trained for 400000 iterations. Xavier initializer was used to initialize weights. Batch size is 64. The network was trained to minimize $l_2$-loss. The convolutions are implemented with no padding. Therefore, for each skip connection, the previous layer's output is cropped to match the input layer. 

For the training process, patches of size $36\time36$ were used. Because we used convolution with no padding, patches get smaller at every layer and the output of the network is $20\times20$. Therefore, while the input patch is $36\times36$ the corresponding label is of size $20\times20$. For each image in the set, as suggested in~\cite{Timofte_2016_CVPR} data augmentation is performed by rotating the image by 90, 180, 270 degrees, flipping and downscaling with 0.9, 0.8, 0.7. This produces 32 image pairs for each training image pair of low (RGB) and corresponding high resolution spectral image. 

\noindent{\textbf{Testing }}
At test time, we use our model without (`ours' setting) and with the enhanced prediction (`ours+E') as suggested in~\cite{Timofte_2016_CVPR}. For the enhanced prediction the input image is rotated and flipped to obtain 8 images processed separately and mapped back to the original state to then average these resulting images for the final result. Generally, using the enhanced prediction is beneficial accuracy-wise (see Table~\ref{tab:MethodComparison}).

\vspace{-0.1cm}
\subsection{Design choices vs. performance}
\label{ssec:subhead}

\begin{figure}[h]
\centering
\vspace{-0.1cm}
\includegraphics[width=\linewidth]{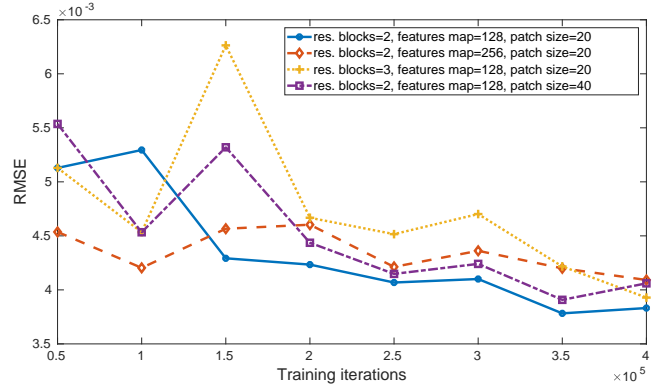}
\vspace{-0.6cm}
\caption{Validation errors (on ICVL) for our method with different design choices.}
\label{fig:design_choices}
\vspace{-0.3cm}
\end{figure}

Figure~\ref{fig:design_choices} shows validation errors for our model with 4 different settings. The number of residual blocks, the number of feature maps and the patch size, respectively,  was varied in the default configuration of our model. For this comparison the ICVL dataset with Canon 1D Mark III response function was used. The dataset was divided into 2 sets and for each set 10 images were set as validation images. For each model, one network is trained on 90 images and tested on corresponding validation images. The results are averaged.
As it can be seen from the figure, after 400,000 iterations the default configuration of our model with 2 residual blocks, 128 features maps and patch size 20 performs the best. Patch size 40 model has a significant problem since training with larger patch size results in a substantial increase in the training time. Moreover, due to memory restrictions, one cannot extract equal number of patches from the training images as the smaller patch size setting which results in higher number of epochs with same iteration number. The runtime is also directly affected by the number of features maps and the number of residual blocks/layers in the model.

\vspace{-0.1cm}
\subsection{Quantitative results}
\label{ssec:quantitative_results}

Table~\ref{tab:MethodComparison} demonstrates the quantitative results of ours and the compared methods. On ICVL and CAVE benchmarks, our method has substantially improved over the competing methods on all metrics. The mean values of the samples in the CAVE dataset are generally lower than those in ICVL, resulting in smaller differences in absolute RMSE. NUS benchmark proved to be more challenging for our network to create the same level of improvement. However, at most metrics our method managed to surpass the state-of-the-art. 

\vspace{-0.1cm}
\subsection{Runtime}
Apart from surpassing the state of the art, due to its shallow architecture, our method is fast. The spectral reconstruction of a RGB image patch of $722\times644$ pixels takes 0.29 seconds on GPU. In order to avoid boundary artifacts, usually, the patches with overlap are given to network at test time. The shallow architecture of our system leads to the ability of operating on larger patches, possibly on whole image, without experiencing RAM issues. This leads to an additional increase in speed of reconstruction.

\vspace{-0.1cm}
\subsection{Visual results}
Figure~\ref{fig:visual_comparison} depicts a qualitative comparison between the reconstruction result at 3 wavelengths achieved by our method and that of A+ for an image from ICVL dataset. For reference we show also the ground truth images. For all 3 wavelengths there is a large and visible improvement, as shown also by the quantitative results on ICVL (Table~\ref{tab:MethodComparison}). 

\vspace{-0.25cm}
\section{Conclusion}
We proposed a novel method for spectral reconstruction from a single RGB image. We avoided overfitting by designing a moderately deep (6 layers) CNN model and careful training. The power of our solution is shown by the relatively low runtime and the state-of-the-art results achieved on the 3 most used spectral reconstruction benchmarks.  
\vspace{-0.2cm}
{
\bibliographystyle{IEEEbib}
\bibliography{strings,refs,references}
}
\end{document}